\documentclass[runningheads]{llncs}
\usepackage{graphicx}
\usepackage{adjustbox}
\usepackage{caption}
\usepackage{environ}
\usepackage{float}
\usepackage[T1]{fontenc}
\usepackage{footnote}
\usepackage{hyperref}
\usepackage{latexsym}
\usepackage{lscape}
\usepackage{microtype}
\usepackage{multicol}
\usepackage{multirow}
\usepackage{nicematrix}
\usepackage[nodisplayskipstretch]{setspace}
\usepackage{rotating}
\usepackage[section]{placeins}
\usepackage{subcaption}
\usepackage{tablefootnote} 
\usepackage{tabularx}
\usepackage{times}
\usepackage{tikz}
\usepackage{url}
\usepackage[utf8]{inputenc}
\usepackage{xspace}

\begin{document}
\title{Document Intelligence Metrics \\ for Visually Rich Document Evaluation}
\author{Jonathan DeGange\inst{1} \and Swapnil Gupta\inst{2} \and
Zhuoyu Han\inst{1} \\ Krzysztof Wilkosz\inst{3} \and Adam Karwan\inst{3}}

\authorrunning{Jonathan DeGange\inst{1} et al.}

\institute{Ernst \& Young (EY) LLP USA \\
\email{\{jonathan.degange, zhuoyu.han\}@ey.com}\and
 EY Global Delivery Services India LLP \\
\email{swapnil.gupta1@gds.ey.com} \and
EY GDS (CS) Poland Sp. z o.o. \\
\email{\{krzysztof.wilkosz, adam.karwan\}@gds.ey.com}}
\maketitle              
\pdfstringdefDisableCommands{%
  \def\\{}%
}

\begin{abstract}
The processing of Visually-Rich Documents (VRDs) is highly important in information extraction tasks associated with Document Intelligence. We introduce \emph{DI-Metrics},  a Python library devoted to VRD model evaluation comprising text-based, geometric-based and hierarchical metrics for information extraction tasks. We apply \emph{DI-Metrics} to evaluate information extraction performance using publicly available \emph{CORD} dataset, comparing performance of three SOTA models and one industry model. The open-source library is available on GitHub\footnote{\url{https://github.com/MetricsDI/DIMetrics}}.

\end{abstract}
\section{Introduction}

Retrieval of the relevant data is often termed Key Information Extraction (KIE) or Information Extraction (IE). Semi-structured forms and documents with complex layout features are commonly known as Visually-Rich Documents (VRD) \cite{liu2019graph}. IE from VRDs is a sub-task of document understanding, often termed Document Intelligence\footnote{\url{https://sites.google.com/view/di2019}} (DI), which applies artificial intelligence and machine learning to business documents and processes. 

Key Information Extraction from VRDs is a challenging task of active research in the research community \cite{Ritesh2021Improving}\cite{Tecuci2020DICR}. Many fields in semi-structured documents such as invoices or receipts are hierarchical (e.g. item description, item count, item total, all roll up to a singular parent line item class), and as previously stated, require two-dimensional processing.  Current SOTA approaches are often based on self-supervised pre-training and transfer learning\footnote{\url{https://docs.microsoft.com/en-us/azure/cognitive-services/form-recognizer}}. Models often comprise a multi-modal representation of the page content's text, location (bounding boxes), and other important visual semantic queues. 

\section{Metrics}
\vspace{-10pt}

We provide a library to ease consistent comparison of VRD model performance on IE tasks. The library is a collection of existing and new IE metrics (Table \ref{table:allMetrics}) accessible through a Python3 API. Many metrics are dynamic programs based on edit distance, and they are known to be computationally expensive. Our implementations are accelerated by pre-compilation in Cython \cite{behnel2011cython}. We also introduce a novel metric for handling evaluation of hierarchical fields, \emph{Unordered Hierarchical Edit Distance} (UHED).
\vspace{-25pt}
\begin{table*}[ht!]
\caption{Metrics available in the the \emph{DI-Metrics} library}
\label{table:allMetrics}
\centering
\begin{tabular}{|l|l|c|}
\hline
\textbf{Metric Type} & \textbf{Metrics Name} & \textbf{Range} \\ 
\hline 
\multirow{3}{6.5em}{\textit{\footnotesize Text-Based\\ (Field Level)}} & Exact Match & \textit{True, False} \\
& Raw Levenshtein Distance & \textit{0 - min(GT, P)} \\ 
& Raw Longest Common Subsequence (LCSeq) & \textit{0 - min(GT, P)} \\
& Token Classification & \textit{0 - 1} \\ %\hline 
\hline 
\multirow{2}{6.5em}{\textit{\footnotesize Geometric-Based (Field Level)}}
& Grouped Bbox by class IoU ($IoU_G$) & \textit{0 - 1} \\
& Constituent Bbox by class IoU ($IoU_C$) & \textit{0 - 1} \\ 
\hline 
\multirow{2}{6.5em}{\textit{\footnotesize Hierarchical (Document Level)}}
& Hierarchical Edit Distance (HED) & \textit{0 - 1} \\
& Unordered Hierarchical Edit Distance (UHED) & \textit{0 - 1} \\ 
\hline
\end{tabular}
\end{table*}
\vspace{-15pt}

Text-based metrics measure the presence of typing or spelling errors and access the convergence of two strings. 

In \textbf{Exact Match (EM)} metric, we simply check whether the entire predicted string \textit{P} is exactly the same as the ground truth string \textit{GT}. 
\textbf{Levenshtein Edit Distance (LED)} between two words is the minimum number of single-character edits (i.e. insertions, deletions or substitutions) required to change one word into the other.
The \textbf{Longest Common Subsequence (LCSeq)} is the minimum number of insertions and deletions required to change one string to the other. 

\vspace{-10pt}
\begin{figure}%
    \centering
    \subfloat[\centering Receipt ground truths (\textcolor{red}{red}), predictions (\textcolor{blue}{blue})]{{\includegraphics[width=4.5cm]{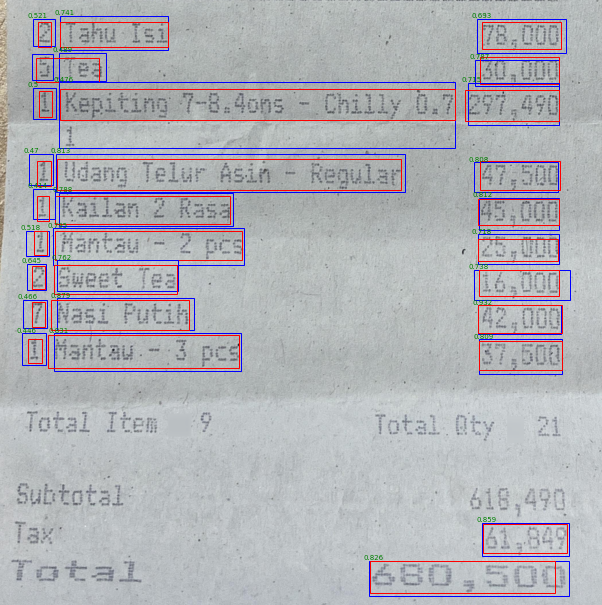}}\label{subfig:gm_a}}
    \qquad
    \subfloat[\centering Geometrical explanation of IoU]{{\includegraphics[width=2.4cm]{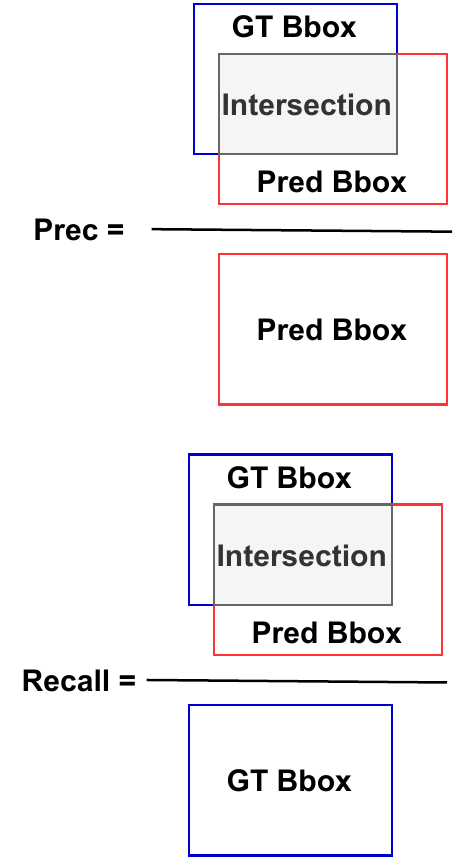}}\label{subfig:gm_b}}
    \vspace{-5pt}
    \caption{Visualization and explanation of geometric-based metrics}
    \label{fig:geometric_metrics}
\end{figure}
\vspace{-10pt}
Geometric-based metrics consider the ratio between the overlap of the location of the area of the text in ground truth and predicted bounding boxes. Geometric-based metrics are useful especially in IE when the targeted text for extraction coincidentally appears in multiple locations on the same page (i.e. right answer, wrong location), and also for document layout analysis tasks.  
Figure~\ref{subfig:gm_a} presents an example of receipt with labeled visualization of ground truth bounding boxes and predicted fields. 
When using \textbf{Grouped Bbox by class} $\textbf{(IoU}_\textbf{G}\textbf{)}$ approach one computes the overlap of aggregated boxes by calculating a convex-hull minimal spanning box of all constituent bounding boxes surrounding the entire field and thus include any spaces between constituent OCR as well.

Similar yet slightly different, \textbf{Constituent Bbox by class} $\textbf{(IoU}_\textbf{C}\textbf{)}$
is adapted from the DocBank dataset paper \cite{LiXu20DocBank}, where instead of taking the area of the entire field, we only consider the areas of individual tokens (words). 

Hierarchical Metrics are applied when the fields of interest are nested. In \cite{Hwang2020Spatial}, edit distances are extended from strings to table cells of strings, using a tree-based edit distance for table cell recognition. \textbf{Hierarchical Edit Distance (HED)} was proposed by \cite{Chua2021DeepCPCFG}. This metric also covers information about non-nested and hierarchical fields (line-items), effectively only requiring that the ordering of line-items within a document and words within a field remain the same, while the ordering of fields within a line-item may be permuted without impacting the distance. Our proposed \textbf{Unordered Hierarchical Edit Distance (UHED)} relaxes HED, allowing unordered lists of line-items. We apply Hungarian assignment algorithm to find the optimal (\textit{GT}, \textit{P}) pairs by minimizing the matrix of input distances for each possible candidate pairs via bipartite matching \cite{jonker1986improving}. 

\vspace{-10pt}
\section{Experimental Results}
\vspace{-10pt}

To test application of the metrics on models and data, we use CORD Receipts dataset. In Table \ref{tab:CORD_HED_UHED} we present a comparison of HED and UHED metrics for three models: LayoutLM Base V1 \cite{xu2020layoutlmv2},  DeepCPCFG \cite{Chua2021DeepCPCFG}, and Microsoft Form Recognizer pre-built receipt model.

\vspace{-25pt}
\begin{table*}
\caption{Ordered and Unordered Hierarchical Edit Distance metrics.} 
\label{tab:CORD_HED_UHED}
\centering
\captionsetup{justification=centering, margin=0.5cm}
\begin{tabular}{lcccccc}
& \multicolumn{3}{c}{\textbf{HED}} &  \multicolumn{3}{c}{\textbf{UHED}} \\
\textbf{CORD} & \textbf{F1\textsuperscript{\textdagger} } & \textbf{Precision\textsuperscript{\textdagger} } & \textbf{Recall\textsuperscript{\textdagger} } & \textbf{F1\textsuperscript{\textdagger} } & \textbf{Precision\textsuperscript{\textdagger} } & \textbf{Recall\textsuperscript{\textdagger} } \\ 
\hline
LayoutLM + PSL LI Rules & 0.89 & 0.88 & 0.91 & 0.92 & 0.92 & 0.94 \\
LayoutLM + Simple LI Rule & 0.86 & 0.85 & 0.89 & 0.92 & 0.96 & 0.90 \\
DeepCPCFG & \textbf{0.96} & \textbf{0.97} & \textbf{0.97} & \textbf{0.97} & \textbf{0.98} & \textbf{0.97} \\
MSFT Form Recognizer & 0.81 & 0.91 & 0.75 & 0.85 & 0.96 & 0.78 \\ \hline
\end{tabular}
\\ \textsuperscript{\textdagger}\footnotesize{Reported values are the mean of F1, precision and recall for each document's HED scores. 

F1 is not directly comparable to precision and recall.}
\end{table*}
\vspace{-10pt}

\textbf{LayoutLM} is a BERT-like transformer model, where bounding box and WordPiece embeddings are summed together as inputs to the transformer hidden layers. We employ sequence labeling approach with single Softmax classifier after the encoder, and train over approximately 18,000 internal proprietary invoices using cross entropy loss function. To group nested line-item classifications, we use Probabilistic Soft Logic (PSL) \cite{Duffy2021DeepPSL} to classify parent line item IDs. The PSL rules combine first-order logic with probabilistic graphical model to perform collective classification  of line-items using outputs from the LayoutLM token classification Softmax classifier. To assess effectiveness of PSL line item grouping, we also implement Simple LI Rule, a rule-based method for assigning bounding boxes group labels. 
\pagebreak 

\textbf{DeepCPCFG} uses an expert-provided grammar and language model potentials as rules, operating on two-dimen\-sio\-nal sequences formed by a directed graph representation of the page structure \cite{Chua2021DeepCPCFG}. Unlike LayoutLM, DeepCPCFG does not require bounding box labels, but uses ground truth key-value pairs as inputs, and latently learns the mapping to bounding boxes on page. 

\textbf{Microsoft Form Recognizer} is used as an industry benchmark end-to-end model, accessible via API calls. We benchmark the pre-built receipt model. We do share results of training a custom model on CORD data, due to inability to create custom parent-child predictions with the API.

\vspace{-10pt}
\section{Discussion and Conclusion}
\vspace{-10pt}
We have shared \emph{DI-Metrics}, a library for objective evaluation of IE Document Intelligence Tasks. The library provides a comprehensive set of metrics for use by researchers and industry practitioners to use and transparently benchmark information extraction models.
In this paper, we also introduced UHED metric.

\textit{Disclaimer:} The views reflected in this article are the views of the authors and do not necessarily reflect the views of the global EY organization or its member firms.

\textit{Acknowledgement:}
The authors would like to thank the following colleagues: Freddy Chua, Sunil Tiyyagura, Hamid Motahari and Nigel Duffy for their thoughtful feedback and suggested edits.

\FloatBarrier
\bibliographystyle{splncs04}
\bibliography{custom}

\begin{thebibliography}{10}
\providecommand{\url}[1]{\texttt{#1}}
\providecommand{\urlprefix}{URL }
\providecommand{\doi}[1]{https://doi.org/#1}

\bibitem{behnel2011cython}
Behnel, S., Bradshaw, R., Citro, C., Dalcin, L., Seljebotn, D.S., Smith, K.:
  Cython: The best of both worlds. Computing in Science \& Engineering
  \textbf{13}(2),  31--39 (2011)

\bibitem{Chua2021DeepCPCFG}
Chua, F.C., Duffy, N.P.: {DeepCPCFG}: Deep learning and context free grammars
  for end-to-end information extraction. In: Llad{\'o}s, J., Lopresti, D.,
  Uchida, S. (eds.) Document Analysis and Recognition -- ICDAR 2021. pp.
  838--853. Springer International Publishing, Cham (2021)

\bibitem{Duffy2021DeepPSL}
Duffy, N.P., Puranam, S.A., Dasaratha, S., Phogat, K.S., Tiyyagura, S.R.:
  {DeepPSL}: End-to-end perception and reasoning with applications to zero shot
  learning (2021)

\bibitem{Hwang2020Spatial}
Hwang, W., Yim, J., Park, S., Yang, S., Seo, M.: Spatial dependency parsing for
  {2D} document understanding. CoRR  \textbf{abs/2005.00642} (2020)

\bibitem{jonker1986improving}
Jonker, R., Volgenant, T.: Improving the {Hungarian} assignment algorithm.
  Operations Research Letters  \textbf{5}(4),  171--175 (1986)

\bibitem{LiXu20DocBank}
Li, M., Xu, Y., Cui, L., Huang, S., Wei, F., Li, Z., Zhou, M.: {DocBank}: {A}
  benchmark dataset for document layout analysis. CoRR  \textbf{abs/2006.01038}
  (2020)

\bibitem{liu2019graph}
Liu, X., Gao, F., Zhang, Q., Zhao, H.: Graph convolution for multimodal
  information extraction from {Visually Rich Documents}. arXiv preprint
  arXiv:1903.11279  (2019)

\bibitem{Ritesh2021Improving}
Sarkhel, R., Nandi, A.: Improving information extraction from {Visually Rich
  Documents} using visual span representations. Proc. VLDB Endow.
  \textbf{14}(5),  822–834 (Jan 2021)

\bibitem{Tecuci2020DICR}
Tecuci, D., Palla, R., Nezhad, H., Ahuja, N., Monteiro, A., Ishkhanov, T.,
  Duffy, N.: {DICR}: {AI} assisted, adaptive platform for contract review.
  Proc. of AAAI  \textbf{34},  13638--13639 (04 2020)

\bibitem{xu2020layoutlmv2}
Xu, Y., Xu, Y., Lv, T., Cui, L., Wei, F., Wang, G., Lu, Y., Florencio, D.,
  Zhang, C., Che, W., Zhang, M., Zhou, L.: {LayoutLMv2}: Multi-modal
  pre-training for {Visually-Rich Document} understanding (2020)

\end{thebibliography}

\end{document}